\documentclass{article}
\usepackage[T1]{fontenc}
\usepackage[utf8]{inputenc}
\usepackage{authblk}
\usepackage{amsmath}
\usepackage{amsfonts}
\usepackage{nicematrix}
\usepackage{arydshln}
\newtheorem{theorem}{Theorem}

\newtheorem{defi}{Definition}
\newtheorem{proof}{Proof}[section]
\usepackage{tabularx}
\usepackage{booktabs}
\usepackage[labelfont=bf, format=plain, justification=raggedright, singlelinecheck=false]{caption}
\usepackage{natbib}
\usepackage{fancybox}
\setcitestyle{numbers, square}
\usepackage{subfigure} 
\usepackage{graphicx} 
\usepackage{float} 
\usepackage[margin=1.5in]{geometry}
\usepackage[english]{babel}
\usepackage{algorithm}
\usepackage[noend]{algpseudocode}
\usepackage{tikz,xcolor,hyperref,tikzscale}
\usetikzlibrary{shapes.geometric, arrows}  
\usetikzlibrary{fit}
\tikzset{highlight/.style={rectangle,
                           fill=blue!15,
                           rounded corners = 0.5 mm,
                           inner sep=1pt,
                           fit=#1}}
\usepackage{verbatim}

\definecolor{lime}{HTML}{A6CE39}

\DeclareRobustCommand{\orcidicon}{
	\begin{tikzpicture}
	\draw[lime, fill=lime] (0,0) 
	circle [radius=0.16] 
	node[white] {{\fontfamily{qag}\selectfont \tiny ID}};
	\draw[white, fill=white] (-0.0625,0.095) 
	circle [radius=0.007];
	\end{tikzpicture}
	\hspace{-2mm}
}
\foreach \x in {A, ..., Z}{\expandafter\xdef\csname orcid\x\endcsname{\noexpand\href{https://orcid.org/\csname orcidauthor\x\endcsname}
		{\noexpand\orcidicon}}
}

\usetikzlibrary{graphs}
\usetikzlibrary{shapes.geometric,fit,positioning,arrows,automata,calc}
\tikzset{
	main/.style={circle, minimum size = 5mm, thick, draw =black!80, node distance = 10mm},
	connect/.style={-latex, thick},
	box/.style={rectangle, draw=black!100}
}
\title{Arithmetical Binary Decision Tree Traversals}
\author{Jinxiong Zhang  \orcidA{}\\
jinxiongzhang@qq.com}
\date{} 
\begin{document}

\maketitle

\begin{abstract}

This paper introduces a series of methodes for traversing binary decision trees using arithmetic operations. 
We present a suite of binary tree traversal algorithms 
that leverage novel representation matrices to flatten the full binary tree structure 
and embed the aggregated internal node Boolean tests into a single binary vector.
Our approach, grounded in maximum inner product search, offers new insights into decision tree.

\end{abstract}
\section{Introduction}

It is important to represent the hierarchical structure information for machine learning.
The recursive regularization \cite{recursiveRegularization} is proposed to hierarchically select features.
The tree regularization is to enhance the interpretability of deep neural networks in \cite{BeyondSparsity}. 
A hierarchical classifier is built from the weights of the softmax layer in convolutional neural networks in \cite{nbdt}.
A generalized tree representation termed \textit{TART} is based on transition matrix in \cite{Tart}. 
The regional tree regularizer is used to optimize the interpretability of deep neural networks in \cite{TreeRegu}.

\textit{QuickScorer}  and \textit{RapidScorer}, as introduced in\cite{lucchese2015quickscorer} and \cite{ye2018rapidscorer}, 
accelerate the additive ensemble of regression trees in learning to rank by utilizing bitvectors to identify false nodes.
The works \cite{lucchese2015quickscorer, dato2016fast, lettich2016gpu, lucchese2017quickscorer, FPGA2021} 
pay more attention to the implementation of additive tree models,
while we focus on the theory of decision tree algorithms.
Although the aim of these works are to speed up the industry-scale tree ensemble models, 
they inspire us to flatten the structure of binary decision trees via bit-matrix,
which is the base to parameterize binary tree traversals.

In the quest for deeper insights of machine learning algorithms, 
we revisit the fundamental concept of decision tree traversals. 
This paper proposes a paradigm shift from traditional logical operations to an arithmetic-based approach, 
streamlining the traversal process and improving algorithmic interpretability.

We begin with the algorithm \ref{alg:QS} originated from \cite{lucchese2015quickscorer} and provide some illustration. 
Then we translate \ref{alg:QS} from the logical operation to the arithmetical operation.
Similarly, we perform an interleaved traversal of soft decision trees and more general trees by means of simple arithmetical operations.
Then, we turn to the evaluation of binary decision trees. 
After exploring the mathematical underpinnings, we will delve into the interpretable representation of our approach. 
In conclusion, we summarize the pivotal contributions and the broader implications of our research, 
highlighting its potential impact on decision tree algorithms.

\section{Tree Traversal Using  Multiplication Operation}

In this section, we will introduce the \textit{QuickScorer} algorithm originated from \cite{lucchese2015quickscorer}
and derive some tree traversal algorithms based on element-wise multiplication operations.

We follow the description of tree model in \cite{lucchese2015quickscorer}
and invite the reader to refer to the original paper for more details.
\begin{quote}
	A decision tree $T(\mathcal{N}, \mathcal{L})$ is composed of internal nodes $\mathcal{N}=\{n_0,n_1,\cdots,n_t\}$ 
	and leaves $\mathcal{L}=\{l_0,l_1,\cdots, l_{t+1}\}$ 
	where each $n\in \mathcal{N}$ is associated with a \textit{Boolean test} and 
	each leaf $l\in \mathcal{L}$ stores the \textit{prediction} $l.val\in\mathbb{R}$.
	All the nodes whose Boolean conditions evaluate to $False$
	are called false nodes, and true nodes otherwise.
	If a visited node in $\mathcal{N}$ is a false one, then the right branch is taken, and the left branch otherwise.
	Indeed, if $n$ is a false node (i.e., its test condition is false), the leaves in the left subtree of $n$ cannot be the exit leaf.
	Similarly, if $n$ is a true node, the leaves in the right subtree of $n$ cannot be the exit leaf.

\end{quote}

\subsection{Tree Traversal Using Bitvectors}

The QuickScorer, a tree-based ranking model, features reduced control hazards, lower branch mis-prediction rates, 
and improved memory access patterns, as initially described in \cite{lucchese2015quickscorer}.
The core of \textit{QuickScorer} is the bitvector representation of the tree nodes.
The exit leaf is identified by the leftmost bit of the bitwise logical $\mathit{AND}$ operations on false nodes' bitvectors.
More implementation or modification can be found in 
\cite{dato2016fast, lettich2016gpu, lucchese2017quickscorer, ye2018rapidscorer, FPGA2021}.
The algorithm \ref{alg:QS} describes the tree traversal in \textit{QuickScorer}\footnote{Note that the nodes of $T$ are numbered in breadth-first order and leaves from left to
right there, which is used to prove the correctness of the algorithm.}.
\begin{algorithm}
	\caption{Tree Traversal Using Bitvectors}\label{alg:QS}
	\begin{algorithmic}[1]
		
		\Statex \textbf{Input}:
		\begin{itemize}
			\item  input feature vector $\mathbf{x}$
			\item  a binary decision tree $T(\mathcal{N, L})$ with
			\begin{itemize}
				\item internal nodes $\mathcal{N}$: $\{n_0,\cdots,n_{t}\}, 1\leq t<\infty$
				\item a set of leaves $\mathcal{L}$: $\{l_0,\cdots, l_{t+1}\}, 1\leq t<\infty$ 
				\item the prediction of leaf $l\in\mathcal{L}$: $l.val\in\mathbb{R}$
				\item node bitvector associated with $n\in \mathcal{N}$: $n.bitvector$
			\end{itemize}
		\end{itemize}
		
		\Statex \textbf{Output}: tree traversal output value
		
		\Procedure{Score}{$\mathbf{x}, T$} 
		\State Initialize the result bitvector as the all-ones vector: $\mathbf{v} \leftarrow \vec{1}$
		\State Find the $Fasle$ nodes $U(\mathbf{x}, T)$ given $\mathbf{x}, T$
		\For{ node $u\in U$}  \Comment{iterate over the false nodes}
		\State $\mathbf{v} \leftarrow \mathbf{v} \wedge u.bitvector $  \Comment{update by bit-wise logical $\mathit{AND}$}
		\EndFor  \label{roy's loop}
		\State $j\leftarrow$ index of the leftmost bit set to 1 of $\mathbf{v}$
		\State return $l_j.val$
		
		\EndProcedure
		
	\end{algorithmic}
\end{algorithm}

Tree traversal in \cite{lucchese2015quickscorer} is based on the \textit{node bitvector}.
Every internal node $n\in\mathcal{N}$ is associated with a  node bitvector denoted by $n.bitvector$ (of the same length), 
acting as a bit-mask that encodes (with 0's) the set of leaves to be removed from the leaf set $\mathcal{L}$ 
when  $n$ is a false node.
\textit{QuickScorer} is described in \cite{ye2018rapidscorer} as follows. 
\begin{quote}
\textit{QuickScorer} maintains a bitvector, composed of $\wedge$ bits, one per leaf, 
to indicate the possible exit leaf candidates with corresponding bits equal to 1. 
\end{quote}

Here we call the bitvector of a node $n\in\mathcal{N}$ in \cite{lucchese2015quickscorer} the left bitvector as defined in \ref{def:left-vec}.
\begin{defi}\label{def:left-vec}
	A node's left bitvector in a decision tree is a binary sequence 
	that marks the leaves in its left  subtree to exclude when the node is evaluated as false. 
	The left matrix $\mathrm{L}$ is composed of these bitvectors as columns for each internal node.
\end{defi}
A left bitvector of a node $n\in\mathcal{N}$ is in $\{0, 1\}^{|\mathcal{N}|+1}$ where
\begin{itemize}
	\item each bit corresponds to a distinct leaf $\mathcal{L}$;
	\item the bit $0$ indicates the left leaf nodes of $n$;
	\item the bit $1$ indicates other leaf nodes.
\end{itemize} 
Each row of the bit-matrix $\mathrm{L}$ corresponds to a leaf.
The left leaves are always assigned $0$ in the bitvector of its predecessor nodes,
so the rightmost leaf always encodes with $1$ in each bitvector in this scheme.


\begin{figure}[h]
	\begin{minipage}[c][][t]{6cm}
		
	\tikzstyle{results}=[rectangle, text centered, draw = black]  
	\tikzstyle{decisions} =[circle, text centered, draw = gray]  
	\tikzstyle{bits} =[rectangle, text centered, draw = white]  
	\tikzstyle{arrow} = [ ->, >=stealth, draw=blue!40]  
	
	\begin{tikzpicture}[node distance=0.4cm]  
	\node[decisions](rootnode){ $n_0$ };  
	
	\node[decisions,below of=rootnode,yshift=-0.4cm,xshift=-1.4cm](node1){ $n_1$};  
	\node[results,below of=node1,yshift=-0.4cm,xshift=-0.5cm](result2){$v_1$};
	\node[results,below of=node1,yshift=-0.4cm,xshift=0.5cm](result3){$v_2$};

	\node[decisions,below of=rootnode,yshift=-0.4cm,xshift=1.4cm](node2){$n_2$};  
	
	\node[decisions,below of=node2,yshift=-0.4cm,xshift=-1.1cm](node3){$n_3$};  
	
	\node[decisions,below of=node2,yshift=-0.4cm,xshift=1.2cm](node4){$n_4$};  
	
	\node[results,below of=node3,yshift=-0.4cm,xshift=-0.5cm](result31){$v_3$};  
	\node[results,below of=node3,yshift=-0.4cm,xshift=0.5cm](result32){$v_4$};  
	\node[results,below of=node4,yshift=-0.4cm,xshift=-0.5cm](result41){$v_5$};  
	\node[results,below of=node4,yshift=-0.4cm,xshift=0.5cm](result42){$v_6$};  
	\draw [arrow] (rootnode) -- node [left,font=\small] { True } (node1);  
	\draw [arrow] (rootnode) -- node [right,font=\small] { False } (node2);  
	\draw [arrow] (node1) -- node [left,font=\small] { True } (result2);  
	\draw [arrow] (node1) -- node [right,font=\small] {  False } (result3);  
	\draw [arrow] (node2) -- node [left,font=\small] { True } (node3);  
	\draw [arrow] (node2) -- node [right,font=\small] { False } (node4);  
	\draw [arrow] (node3) -- node [left,font=\small] { True  } (result31);  
	\draw [arrow] (node3) -- node [right,font=\small] {False  } (result32);  
	\draw [arrow] (node4) -- node [left,font=\small] { True } (result41);  
	\draw [arrow] (node4) -- node [right,font=\small] {  False  } (result42);  
	\end{tikzpicture}
	\end{minipage}
	$\mathrm{L}=\begin{pmatrix}
		0 & 0 & 1 & 1 & 1\\
		0 & 1 & 1 & 1 & 1\\
		1 & 1 & 0 & 0 & 1\\
		1 & 1 & 0 & 1 & 1\\
		1 & 1 & 1 & 1 & 0\\
		1 & 1 & 1 & 1 & 1\\
	\end{pmatrix}$
	$\mathrm{R}=\begin{pmatrix}
		1 & 1 & 1 & 1 & 1\\
		1 & 0 & 1 & 1 & 1\\
		0 & 1 & 1 & 1 & 1\\
		0 & 1 & 1 & 0 & 1\\
		0 & 1 & 0 & 1 & 1\\
		0 & 1 & 0 & 1 & 0\\
	\end{pmatrix}$
	\caption{Visual representation of a binary decision tree(the left),  its left matrix (the middle), and its right matrix (the right),
	illustrating the transformation from graphical tree structure to matrix form.}
\label{fig:ss}
\end{figure}
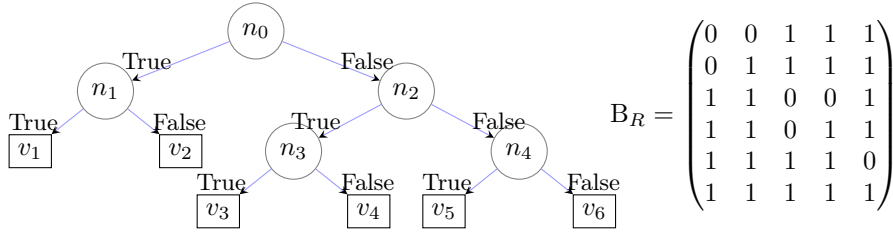
	
Now we consider two simplest cases of tree traversal using left vetors. 
If there is no false nodes for some sample, 
the result bitvector $v$ is only initialized in algorithm \ref{alg:QS}. 
Thus, the first bit must correspond to the leftmost leaf.
If there is no true nodes, 
the result bitvector $v$ is bit-wise logical $\mathit{AND}$ of all the bitvectors,
where the rightmost leaf is indicated by $1$.


In contrast to \cite{lucchese2015quickscorer}, we can define the bitvector 
acting as a bit-mask that encodes (with 0's) the set of leaves to be removed
when the internal node is a true node as following \eqref{def:right-vec}.
\begin{defi}\label{def:right-vec}
	A node's right bitvector in a decision tree is a binary sequence 
	that marks the leaves in its right subtree to exclude when the node is evaluated as true. 	
	The right matrix $\mathrm{R}$ of a decision tree is composed of these bitvectors as columns for each internal node.
\end{defi}
The right bitvector of a node $n\in\mathcal{N}$ is in $\{0, 1\}^{|\mathcal{N}|+1}$, where
\begin{itemize}
	\item each bit corresponds to a distinct leaf $l\in\mathcal{L}$;
	\item the bit $0$ indicates the right leaf nodes of $n$;
	\item the bit $1$ indicates other leaf nodes.
\end{itemize} 
The bitwise logical $\mathit{AND}$  between $\vec{1}$
and the right bitvector of a  node $n$ is to remove the leaves in the right subtree of $n$ from the candidate exit leaves.
Similar to the proof of \ref{alg:QS} in \cite{lucchese2015quickscorer}, we can prove that
the exit leaf is identified by the rightmost bit of the bitwise logical $\mathit{AND}$ on true nodes' right bitvectors.
In another word, the exit leaf is always the one associated with the largest identifier.
And the leftmost leaf always encodes with $1$ in the right bitvectors.

It is able to use both types of bitvectors to perfrom traversals of binary decision trees as shown in algorithm \ref{alg:both}.
We can verify  its correctness by contradiction as similar to the proof of \ref{alg:QS} in \cite{lucchese2015quickscorer}.

\begin{algorithm}[htbp]
	\caption{Tree Traversal Algorithm with Right Vectors and Left Vectors}
	\label{alg:both}
	\begin{algorithmic}[1]
		
		\Procedure{Traversal}{$\mathbf{x}, T(\mathcal{N}, \mathcal{L})$} 
		\State Initialize the result bitvector as the all-ones vector: $\mathbf{v} \leftarrow \vec{1}$ 
		\For{ node $n\in \mathcal{N}$}  
			\If{ node $n$ is a false node}
				\State $\mathbf{v} \leftarrow \mathbf{v} \wedge n.leftvector$ \Comment{update by bitwise logical $\mathit{AND}$}
			\Else{}
				\State $\mathbf{v} \leftarrow \mathbf{v} \wedge  n.rightvector$ 
			\EndIf
			\If{ there is only one  bit set to $1$ in $\mathbf{v}$}
				\State $j\leftarrow$ index of the only bit set to 1 of $\mathbf{v}$
				\State return $l_j.val$

			\EndIf  
		\EndFor    
		
		\EndProcedure
		
	\end{algorithmic}
\end{algorithm}


If the sum of two bitvectors $b$ and $\bar{b}$ is equal to $\vec{1}$, 
they are called orthogonal denoted by $b\bot\bar{b}$.
If the sum of two bit matrices $\mathrm{B}$ and $\mathrm{\bar{B}}$ is equal to a matrix filled with ones,
we say they are complements of each other. 
The bit $1$ indicates the left leaf nodes  in the complements of left matrices
and the right leaf nodes in the complements of right matrices.

We can prove the augmented matrix $(\mathrm{R} \, \vec{1})$ as well as $(\mathrm{L} \, \vec{1})$ is of full rank 
by subtracting the vector $\vec{1}$ from each column in the left matrix $\mathrm{R}$.
Thus, the left bitvectors are linearly independent as well as the right bitvectors and the complements of both.
The rank is really an invariant property of the binary tree's bit-matrix. 

We can prune the decision tree by setting the associated bitvectors to all-zeroes vectors.
It is to prune the second leftmost leaf node in \ref{fig:ss} by setting the second row of the left matrix of \ref{fig:ss} to all-zeroes as the right matrix shown below.
$$
\begin{pNiceMatrix}
	\CodeBefore [create-cell-nodes]
   \Body
	1 & 1 & 1 & 1 & 1\\
	1 & 0 & 1 & 1 & 1\\
	0 & 1 & 1 & 1 & 1\\
	0 & 1 & 1 & 0 & 1\\
	0 & 1 & 0 & 1 & 1\\
	0 & 1 & 0 & 1 & 0\\
	\CodeAfter 
	\tikz \node [highlight = (1-1) (1-5)] {} ;
	\tikz \node [highlight = (2-1) (2-5)] {} ;
	\tikz \node [highlight = (1-1) (6-1)] {} ;
	\tikz \node [highlight = (1-2) (6-2)] {} ;
\end{pNiceMatrix},
\begin{pmatrix}
1 & 1 & 1 & 1 & 1\\
0 & 0 & 0 & 0 & 0\\
0 & 1 & 1 & 1 & 1\\
0 & 1 & 1 & 0 & 1\\
0 & 1 & 0 & 1 & 1\\
0 & 1 & 0 & 1 & 0\\
\end{pmatrix}.$$
The rank of a binary tree's bit-matrix is proper to measure the complexity of the tree.

We embed  the hierarchical structure of the full binary tree into bit-matrices.
It is easy to recursively generate the bit-matrices of perfect binary trees according to their replicated patterns.
The structure  of a decision tree is determined by its bit-matrices. 
It is simple to recovery the structure of a binary decision tree if we have known its left or right bit-matrix.

\subsection{Fuzzy Tree Traversal}

If the test of a visited node in $\mathcal{N}$ is fuzzy, 
then it is to take the right branch or the left branch with probabilities $p\in[0, 1]$ and $1-p$ respectively.
Similarly, we can assign a vector to each internal node, which represents right leaves as $p$ and left leaves as $1-p$.
\begin{defi}\label{def:fuzzy}
	The fuzzy matrix $\mathrm{F}$ of a binary decision tree is a $|\mathcal{L}| \times |\mathcal{N}|$ matrix, 
	where each entry ranges from $0$ to $1$. 
	It represents the probability distribution across leaf nodes, 
	with each node's fuzzy vector as a column. 
	The value $p_n$ denotes the probability of the left leaves, 
	$1-p_n$ for the right leaves, and $1$ for all other leaves.
\end{defi}
And the distribution over the leaf is equal to the product of internal nodes' vectors.

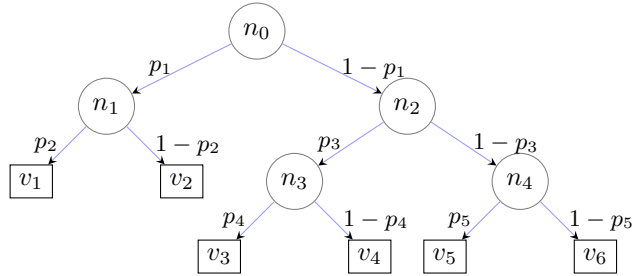
\begin{figure}[h]
\centering
\begin{minipage}[c][][c]{9cm}

	\tikzstyle{results}=[rectangle, text centered, draw = black]  
	\tikzstyle{decisions} =[circle, text centered, draw = gray]  
	\tikzstyle{bits} =[rectangle, text centered, draw = white]  
	\tikzstyle{arrow} = [ ->, >=stealth, draw=blue!40]  
	\begin{tikzpicture}[node distance=0.6cm]  

		\node[decisions](rootnode){ $n_0$ };  

		\node[decisions,below of=rootnode,yshift=-0.4cm,xshift=-2cm](node1){ $n_1$};  
		\node[results,below of=node1,yshift=-0.4cm,xshift=-1cm](result2){$v_1$};
		\node[results,below of=node1,yshift=-0.4cm,xshift=1cm](result3){$v_2$};

		\node[decisions,below of=rootnode,yshift=-0.4cm,xshift=2cm](node2){$n_2$};  

		\node[decisions,below of=node2,yshift=-0.4cm,xshift=-1.5cm](node3){$n_3$};  

		\node[decisions,below of=node2,yshift=-0.4cm,xshift=1.5cm](node4){$n_4$};  

		\node[results,below of=node3,yshift=-0.4cm,xshift=-1cm](result31){$v_3$};  
		\node[results,below of=node3,yshift=-0.4cm,xshift=1cm](result32){$v_4$};  
		\node[results,below of=node4,yshift=-0.4cm,xshift=-1cm](result41){$v_5$};  
		\node[results,below of=node4,yshift=-0.4cm,xshift=1cm](result42){$v_6$};  
		\draw [arrow] (rootnode) -- node [left,font=\small] { $p_1$} (node1);  
		\draw [arrow] (rootnode) -- node [right,font=\small] {$1-p_1$ } (node2);  
		\draw [arrow] (node1) -- node [left,font=\small] { $p_2$ } (result2);  
		\draw [arrow] (node1) -- node [right,font=\small] { $1-p_2$ } (result3);  
		\draw [arrow] (node2) -- node [left,font=\small] {$p_3$ } (node3);  
		\draw [arrow] (node2) -- node [right,font=\small] { $1-p_3$} (node4);  
		\draw [arrow] (node3) -- node [left,font=\small] { $p_4$ } (result31);  
		\draw [arrow] (node3) -- node [right,font=\small] { $1-p_4$ } (result32);  
		\draw [arrow] (node4) -- node [left,font=\small] { $p_5$ } (result41);  
		\draw [arrow] (node4) -- node [right,font=\small] {  $1-p_5$  } (result42); 
	\end{tikzpicture}  
\end{minipage}
\caption{A fuzzy binary decision tree.}
\label{fig:soft} 
\end{figure}

The fuzzy matrix $\mathrm{F}$ of the binary decision tree \ref{fig:soft} has one column for each internalnode of the 
tree and one row for each leaf node of the tree:
$$\mathrm{F}=
\begin{pmatrix}
	p_1 & p_2   & 1 	& 1 	& 1\\
	p_1 & 1-p_2 & 1 	& 1 	& 1\\
  1-p_1 & 1	    & p_3   & p_4 	& 1\\
  1-p_1 & 1 	& p_3   & 1-p_4 & 1\\
  1-p_1 & 1 	& 1-p_3 & 1 	& p_5\\
  1-p_1 & 1 	& 1-p_3 & 1 	& 1-p_5\\
\end{pmatrix}.$$
We can find that the $\mathrm{F}$ is a weighted mixture of the right matrix and  left matrix:
$$\mathrm{F}=
\mathrm{R}\operatorname{diag}(\mathbf{p}) +
\mathrm{L}\operatorname{diag}(\vec{1}- \mathbf{p}),
$$
where the matrix $\mathrm{R}$ and $\mathrm{L}$  is  the right matrix and 
left matrix of the binary decision tree\ref{fig:ss}, respectively;
the vector $\mathbf{p}$ is $(p_1, p_2, p_3, p_4, p_5)^T$; 
and operator $\operatorname{diag}(\mathbf{p})$ 
returns a matrix with diagonal elements specified by the vector $\mathbf{p}$ as described below:
$$
\operatorname{diag}(\mathbf{p})=	
\begin{pmatrix}
p_1 & 0     & 0   & 0 	& 0\\
	0 & p_2 & 0   & 0 	& 0\\
	0 & 0	& p_3 & 0	& 0\\
	0 & 0 	& 0   & p_4 & 0\\
	0 & 0 	& 0   & 0 	& p_5\
\end{pmatrix},
\operatorname{diag}(\vec{1}- \mathbf{p})= 
 \begin{pmatrix}
	1-p_1 & 0     & 0     & 0 	& 0\\
		0 &1- p_2 & 0     & 0 	& 0\\
		0 & 0	  & 1-p_3 & 0	& 0\\
		0 & 0 	  & 0     & 1-p_4 & 0\\
		0 & 0 	  & 0     & 0 	& 1-p_5
\end{pmatrix}.
$$
In general, we can derive the fuzzy matrix $\mathrm{F}$ from the corresponding binary decision trees:
\begin{equation}\label{def:F}
	\mathrm{F}=\mathrm{R}\operatorname{diag}(\mathbf{p})+\mathrm{L}\operatorname{diag}(\vec{1}-\mathbf{p})
\end{equation}
where the vector $\mathbf{p}\in[0, 1]^{|\mathcal{N}|}$ consists of the fuzzy results of branching tests.
The probability on the $i$th leaf, denoted by $p_i$, is the element-wise multiplication of the $i$th row of $\mathrm{F}$:
\begin{equation}\label{eq:leaf}
	p_i=\prod_{j}^{|\mathcal{N}|}\mathrm{F}(i, j),
\end{equation}
where $\mathrm{F}(i, j)$ is the $i$-th element in  the $j$-th column of $\mathrm{F}$.

Specially,  the formula \ref{prob} generates a probability distribution over the leaf nodes 
if every entry of matrix $\mathrm{F}$ is positive.
\begin{equation}\label{prob}
	\mathbf{p}=\exp(\ln(\mathrm{F})\vec{1})
\end{equation}

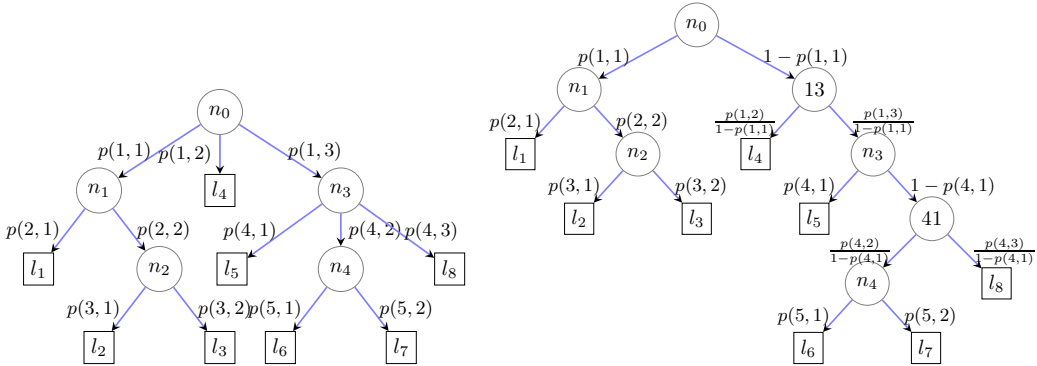
\begin{figure}[htb]
	\tikzstyle{results}=[rectangle, text centered, draw=black]  
	\tikzstyle{decisions} =[circle, text centered, draw = gray]  
	\tikzstyle{bits} =[rectangle, text centered, draw = white]  
	\tikzstyle{arrow} = [thick, ->, >=stealth, draw=blue!50] 
    \begin{minipage}[t]{.45\textwidth}
        \centering

		\resizebox{\columnwidth}{!}{%
			\begin{tikzpicture}[node distance=0.8cm]  
				\node[decisions](rootnode){ $n_1$ };  
		
				\node[decisions,below of=rootnode,yshift=-0.5cm,xshift=-2cm](node1){ $n_2$}; 
				\node[results,below of=rootnode,yshift=-0.5cm, xshift=0cm](result4){$l_5$};  
				\node[decisions,below of=rootnode,yshift=-0.5cm,xshift=2cm](node3){$n_4$};  
		
				\node[results,below of=node1,yshift=-0.5cm, xshift=-1cm](result1){$l_1$};
				\node[decisions,below of=node1,yshift=-0.5cm,xshift=1cm](node2){ $n_2$};  
		
				\node[results,below of=node2,yshift=-0.5cm, xshift=-1cm](result2){$l_2$};
				\node[results,below of=node2,yshift=-0.5cm, xshift=1cm](result3){$l_3$};
		
				\node[results,below of=node3,yshift=-0.5cm,xshift=-1.8cm](result5){$l_5$}; 
				\node[decisions,below of=node3,yshift=-0.5cm,xshift=0cm](node4){$n_5$};  
				\node[results,below of=node3,yshift=-0.5cm,xshift=1.8cm](result8){$l_8$};

				\node[results,below of=node4,yshift=-0.5cm,xshift=-1cm](result6){$l_6$};  
				\node[results,below of=node4,yshift=-0.5cm,xshift=1cm](result7){$l_7$};  
				\draw [arrow] (rootnode) -- node [left,font=\small] { $p(1,1)$ } (node1);  
				\draw [arrow] (rootnode) -- node [left,font=\small] {$p(1,2)$ } (result4);  
				\draw [arrow] (rootnode) -- node [right,font=\small] { $p(1, 3)$ } (node3);  
				\draw [arrow] (node1) -- node [left,font=\small] { $p(2, 1)$  } (result1);  
				\draw [arrow] (node1) -- node [right,font=\small] { $p(2, 2)$ } (node2);  
				\draw [arrow] (node2) -- node [left,font=\small] { $p(3, 1)$ } (result2); 
				\draw [arrow] (node2) -- node [right,font=\small] {$p(3,2)$ } (result3);  
				\draw [arrow] (node3) -- node [left,font=\small] { $p(4, 1)$  } (result5);  
				\draw [arrow] (node3) -- node [right,font=\small] { $p(4, 2)$ } (node4);  
				\draw [arrow] (node3) -- node [right,font=\small] { $p(4, 3)$ } (result8);  
		
				\draw [arrow] (node4) -- node [left,font=\small] {  $p(5, 1)$ } (result6);  
				\draw [arrow] (node4) -- node [right,font=\small] { $p(5, 2)$ } (result7); 
			\end{tikzpicture}  
			}
    \end{minipage}
    \hfill
    \begin{minipage}[b]{0.54\textwidth}
        \centering
		\resizebox{\columnwidth}{!}{%
		\begin{tikzpicture}[node distance=0.6cm]  
			\node[decisions](rootnode){ $n_0$ };  
		
			\node[decisions,below of=rootnode,yshift=-0.5cm,xshift=-2cm](node1){ $n_1$}; 
			\node[decisions,below of=rootnode,yshift=-0.5cm,xshift=2cm](node13){ $i_3$};

			\node[results,below of=node13,yshift=-0.5cm, xshift=-1cm](result4){$l_4$};  
		
			\node[decisions,below of=node13,yshift=-0.5cm,xshift=1cm](node3){$n_3$};  
		
			\node[results,below of=node1,yshift=-0.5cm, xshift=-1cm](result1){$l_1$};
			\node[decisions,below of=node1,yshift=-0.5cm,xshift=1cm](node2){ $n_2$};  
		
			\node[results,below of=node2,yshift=-0.5cm, xshift=-1cm](result2){$l_2$};
			\node[results,below of=node2,yshift=-0.5cm, xshift=1cm](result3){$l_3$};
		
			\node[results,below of=node3,yshift=-0.5cm,xshift=-1cm](result5){$l_5$}; 
			\node[decisions,below of=node3,yshift=-0.5cm,xshift=1cm](node41){$i_1$};  
		
			\node[decisions,below of=node41,yshift=-0.5cm,xshift=-1.1cm](node4){$n_4$};  
			\node[results,below of=node41,yshift=-0.5cm,xshift=1.1cm](result8){$l_8$};

			\node[results,below of=node4,yshift=-0.5cm,xshift=-1cm](result6){$l_6$};  
			\node[results,below of=node4,yshift=-0.5cm,xshift=1cm](result7){$l_7$};  
			\draw [arrow] (rootnode) -- node [left,font=\small] { $p(1,1)$ } (node1); 
			\draw [arrow] (rootnode) -- node [right,font=\small] { $1-p(1,1)$ } (node13);  
		 
			\draw [arrow] (node13) -- node [left,font=\small] {$\frac{p(1,2)}{1-p(1,1)}$ } (result4);  
			\draw [arrow] (node13) -- node [right,font=\small] { $\frac{p(1,3)}{1-p(1,1)}$ } (node3);  
			\draw [arrow] (node1) -- node [left,font=\small] { $p(2, 1)$  } (result1);  
			\draw [arrow] (node1) -- node [right,font=\small] { $p(2, 2)$ } (node2);  
			\draw [arrow] (node2) -- node [left,font=\small] { $p(3, 1)$ } (result2); 
			\draw [arrow] (node2) -- node [right,font=\small] {$p(3,2)$ } (result3);  
			\draw [arrow] (node3) -- node [left,font=\small] { $p(4, 1)$  } (result5);  
			\draw [arrow] (node3) -- node [right,font=\small] { $1-p(4, 1)$  } (node41);  
		
			\draw [arrow] (node41) -- node [left,font=\small] { $\frac{p(4,2)}{1-p(4,1)}$ } (node4);  
			\draw [arrow] (node41) -- node [right,font=\small] {  $\frac{p(4,3)}{1-p(4,1)}$ } (result8);  
		
			\draw [arrow] (node4) -- node [left,font=\small] {  $p(5, 1)$ } (result6);  
			\draw [arrow] (node4) -- node [right,font=\small] { $p(5, 2)$ } (result7); 
		\end{tikzpicture} 
		} 
    \end{minipage}  
    \caption{A  general tree (the left) and its equivalent binary tree (the right) with the same probability distribution over the leaf nodes.}
    \label{fig:any}
\end{figure}

The formula \ref{eq:leaf} is also suitable for general trees.
There are two strategies to obtain the probability on the leaf nodes of  general trees.
The first is to  convert the general tree to a binary tree, 
and then compute the probability on the leaf nodes using the formula \ref{eq:leaf}.
Figure \ref{fig:any} shows a simple general tree and its converted binary tree.
The second is to compute the probability on the leaf nodes using the formula \ref{eq:leaf} directly.
For example, the distribution over the leaves of tree is 
computed by $\exp(\ln(\mathrm{F}\vec{1}))$, where
\begin{equation}\label{eq:any}
\mathrm{F}=\bordermatrix{%
    &  n_1    & n_2 	& n_3 	  & n_4     & n_5 	  \cr
l_1 & p(1, 1) & p(2, 1) & 1 	  & 1 	    & 1	      \cr
l_2 & p(1, 1) & p(2, 2) & p(3, 1) & 1 	    & 1       \cr
l_3 & p(1, 1) & p(2, 2) & p(3, 2) & 1 		& 1       \cr
l_4 & p(1, 2) & 1	  	& 1	      & 1 		& 1       \cr
l_5 & p(1, 3) & 1 	    & 1 	  & p(4, 1) & 1   	  \cr
l_6 & p(1, 3) & 1	    & 1 	  & p(4, 2) & p(5, 1) \cr
l_7 & p(1, 3) & 1 	    & 1 	  & p(4, 2) & p(5, 2) \cr
l_8 & p(1, 3) & 1 	    & 1 	  & p(4, 3) & 1 	\cr
}.\end{equation}
And the number $p(i, j)$ is non-negative for all $i, j$;
$\sum_{j}p(i, j)=1 \quad i\in(1, 2, 3, 4, 5)$.
The probability $p(i, j)$ is dependent on the indexes of internal node and edge.
Each row of \ref{eq:any} is associated with the path from the root to a leaf node, 
and each column of \ref{eq:any} is associated with an internal node in the tree \ref{fig:any}.
If we set $p(1,1)=p(1,2)=0$ and $p(1, 3)=1$, 
it means that we choose the path from the root node $n_0$ to the $l_3$ leaf.


The general rooted tree $GT(\mathcal{N}, \mathcal{L})$ is a directed acyclic graph 
composed of internal node set $\mathcal{N}$ and leaf set $\mathcal{L}$,
where each node in $\mathcal{N}$  directs to at least two child nodes;
every leaf in $\mathcal{L}$ is directed to null.
The directed path in the tree from the root to each leaf node is unique.
From the perspective of representation, 
we can define the path matrix of a general tree $GT(\mathcal{N}, \mathcal{L})$ as described in \ref{def:list}.
\begin{defi}\label{def:list}
	The path matrix for a general tree $GT(\mathcal{N}, \mathcal{L})$ assigns a column to each internal node and a row to each leaf node. 
	Each entry $p(n,l)$, ranging from $0$ to $1$, signifies the weight of the path from the root to the leaf. 
	The sum of weights for paths containing the node $n$ equals $1$, and $1$ represents all other entries.
\end{defi}
There are two types of entries in the path matrix:
\begin{itemize}
	\item the number $p(n, l)\in[0, 1]$ indicates the weight that the internal node $n$ assigns to the path from the root to the leaf $l$;
	\item and number $1$ indicates the rest of entries.
\end{itemize} 
We adapt the definition \ref{def:left-vec} and \ref{def:right-vec} to general trees in \ref{def:mask},
which can mask the inaccessible leaf nodes respect with the internal nodes and their edges.
\begin{defi}\label{def:mask}
For an edge $e$ extending from node $n$, the column bitvector $\vec{b}(n, e)$ has a $0$ for each leaf node 
$l$ connected to  $n$ via other edges, and $1$ for all other entries in the column.
\end{defi}

We can depart the internal nodes' multiple choices 
via the lens of matrix decomposition.
For example, we can represent the matrix $\mathrm{B}$ of \ref{fig:any} as follows:
$$\mathrm{F}=\begin{pmatrix}  
 p(1, 1) & p(2, 1) & 1 	    & 1 	    & 1	      		\\
 p(1, 1) & p(2, 2) & 1 	    & 1 	    & 1       		\\
 p(1, 1) & p(2, 2) & p(3, 1)& 1 		& 1        		\\
 p(1, 2) & 1	   & p(3, 2)& 1 		& 1        		\\
 p(1, 3) & 1 	   & 1 		& p(4, 1)   & 1   	   		\\
 p(1, 3) & 1	   & 1 		& p(4, 2)	& p(5, 1) 	   	\\
 p(1, 3) & 1 	   & 1 		& p(4, 2) 	& p(5, 2)  		\\
 p(1, 3) & 1 	   & 1		& p(4, 3)	& 1 			\\
\end{pmatrix}=
\begin{pmatrix}
	1 & 1 & 1 & 1 & 1  \\
	1 & 0 & 1 & 1 & 1  \\
	1 & 0 & 1 & 1 & 1  \\
	0 & 1 & 0 & 1 & 1  \\
	0 & 1 & 1 & 1 & 1  \\
	0 & 1 & 1 & 0 & 1  \\
	0 & 1 & 0 & 0 & 0  \\
	0 & 1 & 0 & 0 & 1  \\
   \end{pmatrix}\operatorname{diag}(\begin{pmatrix}
	p(1, 1)\\
	p(2, 1)\\
	p(3, 1)\\
	p(4, 1)\\
	p(5, 1)
   \end{pmatrix})+
$$
$$
\begin{pmatrix}
	0 & 0 & 1 & 1 & 1  \\
	0 & 1 & 1 & 1 & 1  \\
	0 & 1 & 0 & 1 & 1  \\
	1 & 1 & 1 & 1 & 1  \\
	0 & 1 & 1 & 0 & 1  \\
	0 & 1 & 1 & 1 & 0  \\
	0 & 1 & 1 & 1 & 1  \\
	0 & 1 & 1 & 0 & 1  \\
   \end{pmatrix}\operatorname{diag}(\begin{pmatrix}
	p(1, 2)\\
	p(2, 2)\\
	p(3, 2)\\
	p(4, 2)\\
	p(5, 2)
   \end{pmatrix})+
   \begin{pmatrix}
	0 & 1 & 1 & 1 & 1  \\
	0 & 1 & 1 & 1 & 1  \\
	0 & 1 & 1 & 1 & 1  \\
	0 & 1 & 1 & 0 & 1  \\
	1 & 1 & 1 & 0 & 1  \\
	1 & 1 & 1 & 0 & 1  \\
	1 & 1 & 1 & 0 & 1  \\
	1 & 1 & 1 & 1 & 1  \\
   \end{pmatrix}\operatorname{diag}(\begin{pmatrix}
	p(1, 3)\\
	0\\
	0\\
	p(4,3)\\
	0
   \end{pmatrix})
$$
where the root nodes $n_0$ and $n_3$ assign the probaility $p(1, 1), p(1, 2), p(1, 3)$ and $p(4, 1), p(4, 2), p(4, 3)$ 
to the their branches respectively.

\section{Decision Tree Evaluation from Diverse Perspectives}

We will convert the bitwise logical $\mathit{AND}$ operations to the arithmetical operations.
Especially, we focus on how to implement binary decision tree in the language of matrix computation. 

\subsection{The Column Perspective}


The first key step in Algorithm \ref{alg:QS} is to  find the false nodes given the input instance and tree.
The second key step in Algorithm \ref{alg:QS} is to find the exit leaf node using bitwise logical $AND$ on these bitvectors. 
In this section, we will implement these steps in the language of matrix computation.


Finding the false nodes in (\ref{alg:QS}) is required to evaluate all the Boolean tests of internal nodes. 
We use a bitvector to represent such evaluation as defined in (\ref{i-vector}).
\begin{defi}\label{i-vector}
	The test vector $\mathbf{t}$ for an input $\mathbf{x}$ is  a column vector
	where  $1$s indicate  false nodes and  $0$s indicate  true nodes in the context of a given tree $T(\mathcal{N}, \mathcal{L})$.
\end{defi}

The bit-wsie logical $\mathit{AND}$ of bitvectors is equivalent to their element-wise multiplication, so 
\begin{equation}\nonumber
\arg\max(\underset{u\in U}{\wedge} u.bitvector)=\arg\max(\underset{u\in U} \prod u.bitvector),
\end{equation}  
where the operator $\arg\max(\cdot)$ returns the indices of all the bits set to $1$ in bitvectors.
If the $i$th bit is set to $1$ in the product of the bitvectors in algorithm (\ref{alg:QS}),
then the $i$th bit of every bitvector is required to be set to $1$.
Thus,  the indices of the bits  set to $1$ in the product of the  bitvectors 
are the arguments of the maximum in their sum as following.
\begin{equation}\label{key}
	\arg\max(\underset{u\in U} \prod u.bitvector)=\underset{u\in U}\cap\arg\max( u.bitvector)=\arg\max(\sum_{u\in U} u.bitvector).
\end{equation}

Based on the test vector $\mathbf{t}$  and left bit-matrix $\mathrm{L}$, 
we can replace the logical $\mathit{AND}$ operation in the algorithm (\ref{alg:QS}) 
using the arithmetical multiplication operation on false nodes' left bitvectors:
\begin{equation}\label{equa:QS}
	\mathbf{v} = \mathrm{L}\mathbf{t} + \vec 1
\end{equation}
where the bitvector $\vec 1$ only consists of the bit set to 1.
And the exit leaf node is identified by the index of the leftmost maximum in $\mathbf{v}$.
As stated above, we have proved the following theorem (\ref{thm:QS}).
\begin{theorem}\label{thm:QS}
	  Algorithm \ref{alg:matrix} identifies the correct exit leaf  for every binary decision tree $T(\mathcal{N}, \mathcal{L})$ and input vector $\mathbf{x}$.
\end{theorem}
\begin{algorithm}
	\caption{Binary decision tree traversal in language of matrix computation}\label{alg:matrix}
	\begin{algorithmic}[1]
		
		\Statex \textbf{Input}:
		\begin{itemize}
			\item  a feature vector $\mathbf{x}$
			\item  a binary decision tree $T(\mathcal{N}, \mathcal{L})$
		\end{itemize}
		
		\Statex \textbf{Output}: tree traversal output value
		\Procedure{Scorer}{$\mathbf{x}, T$} 
		\State Compute the test vector  $\mathbf{t}$ given $\mathbf{x}, T(\mathcal{N}, \mathcal{L})$ \Comment{as defined in \ref{i-vector}}
		\State Compute the result vector: $\mathbf{v}\leftarrow \mathrm{L}\mathbf{t} +\vec{1}$  \Comment{tree traversal via right matrix}
		\State $j\leftarrow  \min\arg\max{\mathbf{v}}$   
				\Comment{identification of exit leaf by maxima}
		\State return $l_j.val$
		
		\EndProcedure
		
	\end{algorithmic}
\end{algorithm} 

Algorithm \ref{alg:matrix} replaces the procedure to find the false nodes 
and the iteration over the false nodes in Algorithm \ref{alg:QS} with the binary embedding of Boolean test evaluatation and matrix-vector multiplication respectively.

If we replace every  $0$ in $\mathrm{L}_i$ by $-1$ to generate a new matrix $\tilde{\mathrm{L}}_i$, then we can find the following relation:
\begin{equation}\label{Eq:negative}
	\arg\max \mathrm{L}\mathbf{t}=\cap_{i}\arg\max(t_i \mathrm{L}_i)
	=\cap_{i}\arg\max(t_i \tilde{\mathrm{L}}_i)
	=\arg\max(\sum_{i}t_i \tilde{\mathrm{L}}_i),
\end{equation}
so  bit-matrix is not the only choice for the algorithm (\ref{alg:QS}). 
The matrix-vector multiplication is the linear combination of column vector,
so $$\mathrm{L}\mathbf{t}=\sum_{i}t_i \mathrm{L}_i,$$
where $\mathrm{L}_i$ is the $i$-th column of the matrix $\mathrm{L}$. 
And we can obtain the following relation:
\begin{equation}\label{Eq:positive}
	\arg\max(\mathrm{L}\mathbf{t} + \vec 1)=\arg\max(\sum_{i}t_i \mathrm{L}_i)=\arg\max(\sum_{i}p_i t_i \mathrm{L}_i)
\end{equation} 
if each $p_i$ is positive.
Thus, the non-negative number is sufficient for the test vector $\mathbf{t}$ and bit-matrix $\mathrm{L}$.
In another word, the domain of Algorithm \ref{alg:matrix} can be extended  from binary values  to non-negative value.


We can follow the same way to convert Algorithm \ref{alg:both} to Algorithm \ref{alg:bothmatrix}.
\begin{algorithm}[htbp]
	\caption{Algorithm \ref{alg:both} in language of matrix computation} \label{alg:bothmatrix}
	\begin{algorithmic}[1]
		\Statex \textbf{Input}:
		\begin{itemize}
			\item  a feature vector $\mathbf{x}$
			\item  a binary decision tree $T(\mathcal{N}, \mathcal{L})$
		\end{itemize}
		
		\Statex \textbf{Output}: tree traversal output value
		\Procedure{Traversal}{$\mathbf{x}, T$} 
		\State Compute the test vector  $\mathbf{t}$ given $\mathbf{x}, T(\mathcal{N}, \mathcal{L})$ \Comment{as defined in \ref{i-vector}}
		\State Compute the result vector: $\mathbf{v}\leftarrow \mathrm{L}\mathbf{t} +\mathrm{R}(\vec{1}-\mathbf{t})$  
		\Comment{as defined in \ref{def:right-vec}  and \ref{def:left-vec} }
		\State $j\leftarrow  \arg\max{\mathbf{v}}$   
				\Comment{identification of exit leaf by maxima}
		\State return $l_j.val$
		
		\EndProcedure
		
	\end{algorithmic}
\end{algorithm}

However, computation procedures may be equivalent in form while different in efficiency.
The $\arg\max(\cdot)$ and intersection operation may take more time to execute.
And these diverse equivalent forms will help us understand the algorithm further.
These observation and facts are the bases to design more efficient tree traversal algorithms.


 The $i$-th value of $\mathrm{L}\mathbf{t}$  is the inner product between the $i$-th row of the $\mathrm{L}$ and the test vector  $\mathbf{t}$, 
 which can imply the number of false nodes that cannot remove the $i$-th candidate leaf node. 
 And the maximum of $\mathrm{L}\mathbf{t}$ is the number of all false nodes.
Next we will introduce another perspective to represent the decision tree traversal in language of matrix computation.

\subsection{The Row Perspective}

\textit{QuickScorer} in \cite{lucchese2015quickscorer} looks like magicians' trick
and its lack of intuitions makes it difficult to understand the deeper insight behind it.  
The bitvector is position-sensitive in \textit{QuickScorer}, 
where the internal nodes and leaves are required to be numbered in breadth-first order and from left to right respectively.

The exit leaf node of the instance $\mathbf{x}$ is determined 
by the internal nodes in the path from the root to the exit leaf node, 
so it is those associated Boolean tests in Algorithm \ref{alg:QS} that play the lead role during traversal rather than the false  bitvectors.
We should take it into consideration when designing new decision tree traversal approaches in language of matrix computation.
For simplicity, we call the path from the root node to a leaf node $l\in\mathcal{L}$ as the decision path of $\ell$ in decision tree $T(\mathcal{N},\mathcal{L})$ thereafter.

\begin{defi}\label{sign-vector}
	The signed test vector $\mathbf{s}$ for an input $\mathbf{x}$ is a binary vector 
	where $1$s signify false nodes and $-1$s signify  true nodes in a decision tree.
\end{defi}
We can find the equation $\mathbf{s}=2\mathbf{t}- \vec 1$ by comparing definition \ref{def:right-vec} and \ref{sign-vector}.

We regard the matrix-vector $\mathrm{L}\mathbf{t}$ in \ref{equa:QS} as the linear combination of bitvectors
while it is the inner products of the row vectors of $\mathrm{L}$ and test vector $\mathbf{t}$ 
that determine the maximum in  $\mathrm{L}\mathbf{t}$.
The bitvectors as columns in matrix $\mathrm{L}$ is designed for the internal nodes rather than the leaf nodes in \ref{alg:QS}.

We need to design vectors from the row perspetive to charaterize the leaf node.
Once the input instance can visit a leaf node $l\in\mathcal{L}$,
the inner product of the $l$'s row vector and the signed test vector is requied to be equal to the number of the internal nodes in $l$'s decision path.

We define the leaf vectors in \ref{leaf-vector}.
And we use inner product to model the Boolean interactions of features of each path in decision trees.
\begin{defi}\label{leaf-vector}
	The leaf vector is a ternary row vector,
	where  $+1$s, $-1$s, $0$s are for false nodes on its decision  path,  true nodes on its decision path,  the rest of internal nodes, respectively.
\end{defi}
We defined the path matrix to describe the structure of the binary decision tree as \ref{signed-matrx}. 
\begin{defi}\label{signed-matrx}
	The  path matrix $\mathrm{P}$ of a decision tree is composed of its leaf vectors as rows.
\end{defi}
For a decision tree, its path matrix $\mathrm{P}$ is equal to the difference of the
left matrix $\mathrm{L}$    and the  right matrix $\mathrm{R}$ or their complement matrices:
\begin{equation}
\mathrm{P}=\mathrm{L}-\mathrm{R}=\bar{\mathrm{R}} - \bar{\mathrm{L}}.
\end{equation}
It is simple to prove that the signed matrix $\mathrm{P}$ is of full column rank by reduction.

It is simple to verify the  following relation on inner products for every $\mathbf{p}\in \{+1, 0, -1\}^d$:
\begin{equation}\label{upper-bound}
	\max_{\mathbf{s}\in\{+1,  -1\}^d}\left<\mathbf{p}, \mathbf{s}\right>=\|\mathbf{p}\|_2^2 \iff \left<\mathbf{p}, \mathbf{s}-\mathbf{p}\right> = 0.
\end{equation}
If $p_is_i$ is equal to $1$,  we say that the leaf node and the input instance reach a consensus on the $i$-th test.
If the input instance can visit a leaf node, they can reach the consensus on the $i$-th test when $p_i\neq 0$.
The depth of the leaf node is the upper bound of those inner products of the row  in $\mathrm{P}$ associated with this leaf  and all possible test vectors.

\begin{defi}\label{condition-number}
	The depth vector $\mathbf{d}$ of a decision tree is a vector containing the depths of its leaf nodes. 
	The depth matrix (denoted $\mathrm{D}$) is defined as a diagonal matrix with these depths.
\end{defi}

If the number of consensuses reached between the input and the leaf node is equal to the depth of the leaf node, 
we prove that the leaf is the exit of the input in the decision tree in \ref{thm:matrix}.

We propose the algorithm termed \textit{SignQuickScorer} as  following.
\begin{algorithm}
	\caption{\textit{SignQuickScorer}}
	\label{alg:signmatrix}
	\begin{algorithmic}[1]
		\Statex \textbf{Input}:
		\begin{itemize}
			\item  a feature vector $\mathbf{x}$
			\item  a binary decision tree $T(\mathcal{N}, \mathcal{L})$
		\end{itemize}
		
		\Statex \textbf{Output}: tree traversal output value
		\Procedure{SignQuickScorer}{$\mathbf{x}, T(\mathcal{N}, \mathcal{L})$} 
		\State Find the signed  test vector  $\mathbf{s}$ given $\mathbf{x}, T$ \Comment{ as defined in \ref{sign-vector}}
		\State Compute the result vector: $\mathbf{v}\leftarrow \mathrm{D}^{-1}\mathrm{P}\mathbf{s}$  
			\Comment{ $\mathrm{D}$ and $\mathrm{P}$  defined in \ref{condition-number} and \ref{signed-matrx} respectively.}
		\State $j\leftarrow \arg\max\mathbf{v}  $  \Comment{index of the maximum of $\mathbf{v}$}
		\State return $l_j.val$
		\EndProcedure	
	\end{algorithmic}
\end{algorithm}

\begin{theorem}\label{thm:matrix}
	Algorithm \ref{alg:signmatrix} identifies the correct exit leaf $e$ for every binary decision tree $T$ and input feature vector $\mathbf{x}$.
\end{theorem}
\begin{proof}
We only need to prove that the unique maximum of $\mathrm{D}^{-1}\mathrm{P}\mathbf{s}$
is the  inner product of the normalized exit  leaf  vector $\frac{s^{\ast}}{\left< s^{\ast}, s^{\ast}\right>}$ and the test vector $\mathbf{s}$,
where $s^{\ast}$ is the representation vector of the exit leaf node.

First, we prove that $\frac{\left<s^{\ast},\mathbf{s}\right>}{\left<s^{\ast},s^{\ast}\right>}=\max\{\mathrm{D}^{-1}\mathrm{P}\mathbf{s}\}$.

According to  the inequality \eqref{upper-bound}, $\max\{\mathrm{D}^{-1}\mathrm{P}\mathbf{s}\}\leq 1$.

According to the definition \ref{sign-vector} and \ref{leaf-vector}, 
we find $\left<s^{\ast},\mathbf{s}-s^{\ast}\right>=0$ 
which implies that $\left<s^{\ast},\mathbf{s}\right>=\left<s^{\ast},s^{\ast}\right>$ 
so $\max\{\mathrm{D}^{-1}\mathrm{P}\mathbf{s}\}=\frac{\left<s^{\ast},\mathbf{s}\right>}{\left<s^{\ast},s^{\ast}\right>}=1$.

Because at least one non-zero elements of the rest of the leaf vector $\mathrm{p}$ are not equal to the coresponding elements in the test vector $\mathbf{s}$, 
we have  $\left<\mathrm{p},\mathbf{s}-s\right>\not=0$   
so that $\left<\mathrm{p}, \mathbf{s}\right> <\left<\mathrm{p}, \mathrm{p}\right>$. 
And we complete this proof.
\end{proof}

And the result vector can be divided element-wisely by the number of the internal nodes over the path, 
then we can obtain the unique maximum as the indicator of the exit leaf.

We can take the example in figure \ref{fig:ss} as follows
$$
\mathbf{s}=\begin{pmatrix}+1\\ +1\\ -1\\ -1\\ +1 \\\end{pmatrix},
\mathrm{P}=\begin{pmatrix}
-1 & -1 & 0 & 0 & 0\\
-1 & +1 & 0 & 0 & 0\\
+1 & 0 & -1 & -1 & 0\\
+1 & 0 & -1 & +1 & 0\\
+1 & 0 & +1 & 0 & -1\\
+1 & 0 & +1 & 0 & +1\\
\end{pmatrix},
\mathrm{D}=\begin{pmatrix}
2 & 0 & 0 & 0 & 0 & 0\\
0 & 2 & 0 & 0 & 0 & 0\\
0 & 0 & 3 & 0 & 0 & 0\\
0 & 0 & 0 & 3 & 0 & 0\\
0 & 0 & 0 & 0 & 3 & 0\\
0 & 0 & 0 & 0 & 0 & 3\\
\end{pmatrix},
\mathrm{D}^{-1}\mathrm{P}\mathbf{s}=\begin{pmatrix}-1\\ 0\\ 1\\ \frac{1}{3}\\ -\frac{1}{3} \\ \frac{1}{3}\end{pmatrix}.
$$

We will show how to take advantage of the unique maximizer of $\mathrm{D}^{-1}\mathrm{P}\mathbf{s}$ in the next subsection.

\subsection{The Algorithmic Perspective}

If we regard each internal nodes as a binary classifier, 
the decision tree is, in essence, of model ensemble. 
If we embed the Boolean tests of raw features into the signed test vector $\mathbf{t}$ in $\{+1, -1\}^d$ in \ref{alg:signmatrix},
it is totally $d+1$ patterns that is determined by the leaf vectors in $\{+1, 0, -1\}^d$,
where different patterns may share the same target value. 
The algorithm(\ref{alg:knn}) regard the algorithm (\ref{alg:signmatrix}) as the decoding of error-correcting output coding 
\cite{escalera2006decoding, escalerasergio2010on}, 
which is initially designed for multi-classification by stacking different binary classifiers.
\begin{algorithm}
	\caption{ \textit{SignQuickScorer} as error-correcting output coding}\label{alg:knn}
	\begin{algorithmic}[1]
		\Statex \textbf{Input}:
		\begin{itemize}
			\item  a feature vector $\mathbf{x}$
			\item  a binary decision tree $T(\mathcal{N}, \mathcal{L})$
		\end{itemize}	
		\Statex \textbf{Output}: tree traversal output value
		\Procedure{Traversal}{$\mathbf{x}, T(\mathcal{N}, \mathcal{L})$} 
		\State Compute the signed test vector  $\mathbf{s}$ given $\mathbf{x}, T$ 
		\Comment{as defined in \ref{sign-vector}}
		\For{ leaf $i\in \mathcal{L}$}  
		\State Compute the similarity between $\mathbf{s}$ and the leaf vector $P_i$: 
		       $m_i=\frac{\left < P_i, \mathbf{s} \right >}{\left< P_i, P_i \right>}$
		\If{ $m_i$ is equal to 1 }
		\State return $l_i.val$
		\EndIf
		\EndFor   
			
		\EndProcedure	
	\end{algorithmic}
\end{algorithm}

Each class is attached to a unique code in error-correcting output coding(ECOC) \cite{dietterich1994solving, Kong95error-correctingoutput, allwein2001reducing} 
while  a leaf vector only indicates a specific pattern of the feature space 
and each class may be attached to more than one leaves in (\ref{alg:signmatrix}).
And the key step in (\ref{alg:knn}) is to find the maximum similarity, 
while key step in (\ref{alg:signmatrix}) is to find the maximum scaled inner product.

The algorithm \ref{alg:knn} is a universal scheme to convert the rule-based method to the attention mechanism. 
We define the $\delta(\cdot)$ function to simplify \ref{alg:signmatrix}:
\begin{equation} \nonumber
\delta(z) =
\begin{cases}
	1 & z=0 \\
0 & z\not=0 
\end{cases} \quad \forall z\in\mathbb{R}.
\end{equation}
According to the proof of \ref{thm:matrix}, 
we can find the argument of the maximum of $\mathbf{v}$ in \ref{alg:signmatrix} as follows:
$$j= \sum_{i=1}^{|\mathcal{L}|}\delta(\mathbf{v}_i-1)i$$
because that $\max \mathbf{v}=1$ and $\arg\max\mathbf{v}$ is unique, 
where $\mathbf{v}=\mathrm{D}^{-1}\mathrm{P}\mathbf{t}$.

Note that $\mathbf{v}_i=1 \iff \frac{\left<S_i, \mathbf{t}\right>}{\mathbf{d}_i}=1\iff \left<S_i, \mathbf{t}\right>=\mathbf{d}_i$ in \ref{alg:signmatrix}
where the vector $S_i$ is the $i$th row of $S$; the element in the vector is indexed by subscript.
So we can obtain a variant of the algorithm (\ref{alg:signmatrix}) as shown in (\ref{alg:depth}).

\begin{algorithm}
	\caption{\textit{SignQuickScorer}}\label{alg:depth}
	\begin{algorithmic}[1]
		\Statex \textbf{Input}:
		\begin{itemize}
			\item  a feature vector $\mathbf{x}$
			\item  a binary decision tree $T(\mathcal{N}, \mathcal{L})$
		\end{itemize}
		\Statex \textbf{Output}: tree traversal output value
		
		\Procedure{SignQuickScorer}{$\mathbf{x}, T(\mathcal{N}, \mathcal{L})$} 
		\State Find the signed test vector  $\mathbf{s}$ given $\mathbf{x}, T(\mathcal{N}, \mathcal{L})$
		\Comment{$\mathbf{s}$ defined in \ref{sign-vector}.}
		\State Compute the result vector: $\mathbf{v}\leftarrow \mathrm{P}\mathbf{s} -\mathbf{d}.$  
		\Comment{$\mathrm{P},\mathbf{d}$ defined in (\ref{signed-matrx}), (\ref{condition-number}), respectively.}
	    \State Return $\sum_{i}^{|\mathcal{L}|}\delta(\mathbf{v}_i )l_i.val$
		\EndProcedure	
	\end{algorithmic}
\end{algorithm}

The Boolean expression may occur in different path 
so that we can rewrite the signed test vector $\mathbf{s}$ as a selection of the raw  Boolean expression $\mathbf{s}=\mathrm{M}\vec{e}$,
where the matrix $\mathrm{M}$ describe the contribution of each Boolean expression and the importance of each attribute.



If each internal (or branching) node in $T$ is associated with a Boolean test of the form $\mathbf{w}_i^T\mathbf{x} > \gamma_i$,
we can compute the signed test vector $\mathbf{s}$ via a popular linear hash functions $\mathbf{s}=\operatorname{sgn}(\mathrm{W}\mathbf{x} - \gamma)$ \cite{l2rwang2028}
where $\operatorname{sgn}(z) = 1$ if $z >0$  and $\operatorname{sgn}(z) = -1$ otherwise.
In this case, the embedding of Boolean tests is to partition the whole raw feature space.
The signed test vector is acting as the hashing code to represent the relative positions with the hyperlines.
And finding the exit leaf nodes is a special case of the nearest neighbor search problem.
The lack of feature transformation may restrict their predictability and stability of the algorithms (\ref{alg:signmatrix}, \ref{alg:knn}). 
We can apply more diverse hashing functions in \cite{l2rwang2028} to extend these algorithms.

The algorithms \ref{alg:matrix}, \ref{alg:signmatrix} and \ref{alg:knn} identify the exit leaf node by the maximum inner product search\cite{pmlr-v48-mussmann16}.
Given a set of points $\mathcal{K}=(k_1, k_2, \cdots, k_m)$ and a query point $q$, the maximum inner product search is to find 
$$\arg\max_{k_i\in\mathcal{K}}\left<k_i, q\right>$$
where $\arg\max(\cdot)$ returns the indices of the maximum values.
In our cases, the query point is the signed or unsigned test vector 
and the set of points is the set of normalized leaf vectors.
We have proved that the maximum value is unique in the algorithms \ref{alg:matrix}, \ref{alg:signmatrix} and \ref{alg:knn}. 
We can take the advantage of the maximum inner product search to accelerate the inference process.

The following formula  generates a smooth probability distribution  over the leaf nodes
without the $\delta(\cdot)$ operator:
$$\operatorname{softmax}(\mathrm{D}^{-1}\mathrm{P}\mathbf{s}).$$
Interpretability is partially dependent on the model complexity.
The sparsity is regarded as a proxy for interpretability for some model.
However, interpretability is independent of the model complexity for decision trees 
because of their explicit decision boundaries. 
Thus, the algorithm \ref{alg:depth} may provide a further understanding of hard attention.


\section{Discussion}

We clarify some properties of the bit-matrix of decision tree in \textit{QuickScorer}. 
And we introduce the emergent recursive patterns in the bit-matrix 
and some novel  tree traversal algorithms, 
which is used to evaluate the binary decision tree in the language of matrix computation.
It  is able to perform decision tree traversal from diverse perspectives.

The maximum inner product search is the most expensive computation of our algorithms such as \ref{alg:matrix} and \ref{alg:signmatrix}.
We plan to find the prototype sample of each leaf node 
and convert the decision problem into the nearest neighbor search problem directly.
Moreover, we aim at investigating the  deep models based on decision trees.
Finally, we plan to apply the algorithms to model more hierarchical structures.

In conclusion, our work presents a transformative approach to decision tree traversals, with potential applications across various domains. We anticipate further exploration into the integration of these methods with deep learning models.

\bibliographystyle{plain}
\bibliography{Traversals}
\end{document}